# QReLU and m-QReLU: Two novel quantum activation functions to aid medical diagnostics

Luca Parisi[1,2,3*], Daniel Neagu[1], Renfei Ma[2,4], Felician Campean[1]

[1]*Faculty of Engineering & Informatics, University of Bradford, Richmond Road, Bradford, BD7 1DP, United Kingdom*
[2]*University of Auckland Rehabilitative Technologies Association (UARTA), University of Auckland, 11 Symonds Street, Auckland, 1010, New Zealand*
[3]*Parkinson's UK, 215 Vauxhall Bridge Rd, Pimlico, London SW1V 1EJ, United Kingdom*
[4]*Warshel Institute for Computational Biology, The Chinese University of Hong Kong, Shenzhen (CUHK−SZ), China*

**Declarations of interest**: none.

**Contributors**: All authors directly participated in the planning, execution, and analysis in the study. All authors also approved the final version of the manuscript, and this submission.

**Funding**: HEIF 2020 University of Bradford COVID-19 response-funded project 'Quantum ReLU-based COVID-19 Detector: A Quantum Activation Function for Deep Learning to Improve Diagnostics and Prognostics of COVID-19 from Non-ionising Medical Imaging'.

---

**\* Corresponding author (L. Parisi).**
*E-mail address*: luca.parisi@ieee.org. *ORCID*: 0000-0002-5865-8708.

*E-mail address*: d.neagu@bradford.ac.uk (D. Neagu). *ORCID*: 0000-0002-7038-106X.
*E-mail address*: marenfei@cuhk.edu.cn (R. Ma). *ORCID*: 0000-0002-2495-4787.
*E-mail address*: f.campean@bradford.ac.uk (F. Campean). *ORCID*: 0000-0003-4166-8077.

**Abstract**—The traditional and classical ReLU activation function (AF) has been extensively applied in deep neural networks, in particular Convolutional Neural Networks (CNN), for image classification despite its unresolved 'dying ReLU problem', which poses challenges to reliable applications of CNN in both academic and industry research and development. This issue leading to misclassifications and lack of generalisation has obvious important implications for critical applications, such as those in healthcare. Recent approaches are still stepping in a similar direction by just proposing variations of the activation function while maintaining the solution within the same unresolved 'dying ReLU' challenge. This contribution reports a different research direction by investigating the development of an innovative quantum approach to the ReLU AF that avoids the 'dying ReLU problem' by disruptive design. The proposed Leaky ReLU approach was leveraged as a baseline ReLU function on which the two quantum principles of entanglement and superposition were applied to derive the proposed Quantum ReLU (QReLU) and the modified-QReLU (m-QReLU) activation functions. Both QReLU and m-QReLU are implemented and made freely available in TensorFlow and Keras. This original approach is effective and validated extensively in case studies that facilitate the detection of COVID-19 and Parkinson's Disease (PD) from medical images. The two novel AFs were evaluated in a two-layered convolutional neural network (CNN) against nine ReLU-based AFs on seven benchmark datasets. The MNIST dataset was used for an initial validation. Thereafter, the CNN with the two proposed Quantum Activation Functions was evaluated on four benchmark datasets on images of spiral drawings taken via graphic tablets from patients with Parkinson's Disease and healthy subjects, as well as two benchmark datasets on point-of-care ultrasound images on the lungs of patients with COVID-19, those with pneumonia and healthy controls. Despite a higher computational cost, results indicated an overall higher classification accuracy, precision, recall and F1-score brought about by either quantum AFs on five of the seven benchmark datasets, thus demonstrating its potential to be the new benchmark or gold standard AF in CNNs and aid image classification tasks involved in critical applications, such as medical diagnoses of COVID-19 and PD.

———————————————— ◆ ————————————————

# 1 INTRODUCTION

## 1.1 Convolutional Neural Networks for image classification in healthcare

The advancements in medical imaging technologies have generated an increasing amount of medical data, thus demanding more complex computational models able to learn from them and leading to a new field of Machine Learning (ML) named 'Deep Learning' (DL) (LeCun *et al.*, 2015). Deep learning models, e.g., Convolutional Neural Networks (CNN) (LeCun & Bengio, 1995), extend the traditional artificial neural networks (ANNs) e.g. Multi-Layer Perceptron (MLP) (Rumelhart *et al.*, 1986), with multiple hidden layers and of different nature, able to perform feature extraction intrinsically (LeCun *et al.*, 1998). Although CNNs typically require less feature engineering than ANNs, appropriate image pre-processing is needed to ensure patterns are extracted correctly and reliably from raw images (Beam & Kohane, 2018).

CNNs can process either two- or three-dimensional images ingested in their input layer, thus making them suitable to learn from medical images in radiology, such as from ionising techniques, e.g., computed tomography (CT) and X-rays, and non-ionising methods too, e.g., magnetic resonance imaging (MRI) and ultrasound (US) (Ker *et al.*, 2017). The hidden layers in a CNN are called convolutional layers from the 'convolutional' filters used in them to create feature maps from the input images, which represent different features extracted from the input images (LeCun *et al.*, 2015). Activation functions (AFs) are an extremely important component influencing the output of a CNN's architecture, as they are applied per each layer and for every neuron of each layer.

Despite the progress made to facilitate learning in CNNs via pre-training (Hinton & Salakhutdinov, 2006), and to expand CNNs in the form of deeper architectures, e.g., AlexNet (Krizhevsky, Sutskever, & Hinton, 2012), VGG16, GoogleNet and ResNet (Russakovsky, et al., 2015), to improve image classification performance on benchmark datasets, such as ImageNet, the activation functions (AFs) used in CNNs and in its derived architectures have remained almost unchanged (Litjens, et al., 2017). For instance, CNNs with three to four convolutional layers (Siddique *et al.*, 2019; Ahlawat *et al.*, 2020), followed by CNN-derived deeper models, such as ResNet and DenseNet (Chen *et al.*, 2018), achieved state-of-the-art performance in classifying the MNIST benchmark data of handwritten digits (LeCun *et al.*, 1998).

Thus, such parallel developments in deeper networks and existing drawback of leveraging non-optimal AFs have led to attempts to apply such deep neural networks for medical diagnostic applications with a varying degree of success. In fact, whilst deep CNNs, whose hyperparameters were optimised via evolutionary algorithms, such as particle swarm optimisation, achieved clinically acceptable classification performance (Pereira *et al.*, 2016c) in facilitating the detection of Parkinson's Disease (PD) by mining patterns from spiral drawings, deeper CNN-derived architectures, such as the sixteen-layered POCOVID-Net model (which builds on the VGG 16 model) attained only recently a satisfactory performance in detecting the COrona VIrus Disease (COVID-19) from bacterial pneumonia based on patterns from US images (Born *et al.*, 2020).

SARS-CoV-2 is responsible for COVID-19, the 'severe acute respiratory syndrome coronavirus 2' (Cohen & Normile, 2020) and the current global pandemic announced by the World Health Organization (WHO, Mar 2020). This virus leads to respiratory disease in humans (Cui *et al.*, 2019), but it may take from 2 to 14 days for the initial symptoms, e.g., fever and cough, to become manifest after an infection (Centers for Disease Control and Prevention, 2020). However, more severe symptoms can progress to viral pneumonia and typically require mechanical ventilation to assist patients with breathing (Verity *et al.*, 2020). In some more severe cases, COVID-19 can also lead to worsen symptoms and even death (Zhou, et al., 2020), as well as it may be an aetiology of PD itself (Beauchamp *et al.*, 2020).

Thus, it is important to be able to detect neurodegenerative co-morbidities in vulnerable undiagnosed patients, such as PD, promptly and non-invasively too, for example via CNNs that can recognise patterns from spiral drawings, and then applying non-ionising medical imaging techniques (Bhaskar *et al.*, 2020), which are more appropriate for such patients, to facilitate a prompt diagnosis of COVID-19 to improve clinical outcomes. Whilst tremors can be detected from patterns in spiral drawings as indicators of early PD, ground-glass opacities, lung consolidation, bilateral patchy shadowing and relevant other lesions-like patterns can be detected as biomarkers to identify COVID-19-related pneumonia from any other types, including both viral and bacterial pneumonia (Shi *et al.*, 2020). Improvements in the AFs of CNNs can help to improve generalisation in both these image classification tasks.

## 1.2 ReLU activation function and the 'dying ReLU' problem

Different layers of a deep neural network represent various degrees of abstraction, thus capturing a varying extent of patterns from input images (Zeiler & Fergus, 2014). AFs provide the CNN with the non-linearity required to learn from non-linearly distributed data, even in presence of a reasonable amount of noise. An AF defines the gradient of a layer, which depends on its domain and the range. AFs are differentiable and can be either saturated or unsaturated. In **Table 1** the main activation functions commonly used in deep neural networks, including the convolutional neural network (CNN), with their equations and references, are summarised, and introduced below.

Saturated AFs are continuous with their outputs threshold into finite boundaries, typically represented as S-shaped curves, also named 'sigmoidal' or 'squashing' AFs, e.g., the logistic sigmoidal function with its output in the range of 0 and 1 (Liew *et al.*, 2016). Saturated AFs are typically applied in shallow neural networks, e.g., in MLPs. However, saturated AFs lead to the vanishing gradient issue whilst training a network with back-propagation (Cui, 2018), i.e., results in gradients that are less than 1, which become smaller with multiple differentiations and ultimately become 0 or 'vanish'. Thus, changes in the activated neurons do not lead to modifications of any weights during back-propagation. Moreover, the exploding gradient problem can occur, which has an opposite effect to vanishing gradients, wherein the error gradient in the weight is so high that it leads to instability whilst updating the weights during back-propagation. Hyperbolic tangent or 'tanh' (see **Table 1**) is a further saturated AF, but it attempts to mitigate this issue by extending the range of the logistic function from -1 to 1, centred at 0. Nevertheless, tanh still does not solve the vanishing gradient problem.

Unsaturated functions are not bounded in any output ranges and are centred at 0. The Rectified Linear Unit (ReLu) (**Table 1**) is the most widely applied unsaturated AF in deep neural networks, e.g., in CNNs, which provides faster convergence than logistic sigmoidal (LeCun *et al.*, 1998) and tanh AFs, as well as improved generalisation (Litjens *et al.*, 2017). In fact, ReLU generally leads to more efficient updates of weights during the back-propagation training process (Gao *et al.*, 2020). The ReLU's gradient (or slope) is either one for positive inputs or zero for negative ones, thus solving the vanishing gradient issue. Nevertheless, despite providing appropriate initialisation of the weights to small random values via the He initialisation stage (Glorot *et al.*, 2011), with large weight updates, the summed input to the ReLU activation function is always negative ('dying ReLU' problem). This negative value yields a zero value at the output and the corresponding nodes do not have any influence on the neural network (Abdelhafiz *et al.*, 2019), which can lead to misclassification resulting in lack of ability in detecting a pathology involved in an image classification task accurately and reliably, such as for COVID-19 or PD diagnostics.

In an attempt to mitigate the 'dying ReLU' issue, in CNNs and deeper CNN-derived networks (e.g., AlexNet, VGG 16, ResNet, etc.), multiple variations of the ReLU AF have been introduced, such as the Leaky ReLU (LReLU), the Parametric ReLU (PReLU), the Randomised ReLU (RReLU) and the Concatenated ReLU (CReLU), as summarised in **Table 1**. Maas *et al.* (2013) introduced Leaky ReLU (LReLU) to provide a small negative gradient for negative inputs into a ReLU function, instead of being 0. A constant variable $\alpha$, with a default value of 0.01, was used to compute the output for negative inputs (**Table 1**). Leaky ReLU is implemented in the two most popular Python open source libraries for DL named 'TensorFlow' and 'Keras' with the default values of $\alpha$ set to 0.2 and 0.3 respectively (TensorFlow Core v2.3.0: tf.nn.leaky_relu, 30) and (TensorFlow Core v2.3.0: tf.keras.layers.LeakyReLU, 30). Via this modification, LReLU leads to small improvements in classification performance overall as compared to the ReLU AF.

Another variant of ReLU, named 'Exponential Linear Unit' (ELU) is aimed at improving convergence (Maas *et al.*, 2013) (**Table 1**), but it still does not solve the 'dying ReLU' issue either. Klambauer *et al.* (2017) introduced a variant of ELU called 'Scaled Exponential linear Unit' (SELU) (**Table 1**), which is a self-normalising function that provides an output as a normal distribution graph, making it suitable for deep neural networks with the output converging to zero mean when passed through multiple layers. Although SELU attempts to avoid both vanishing and exploding gradient problems, it does not mitigate the 'dying ReLU' issue.

He *et al.* (2015) proposed the Parametric Rectified Linear Unit (PReLU) in an attempt to provide a better performance than ReLU in large-scale image classification tasks, although the only difference from LReLU is that $\alpha$ is not a constant and it is learned during training via back-propagation. Nevertheless, due to this, the PReLU does not solve the 'dying ReLU' issue either, as it is intrinsically a slight variation of the LReLU AF. Similarly, the Randomised Leaky Rectified Linear Unit is a randomised version of LReLU (Pedamonti, 2018), whereby $\alpha$ is a random number sampled from a uniform distribution, thus being still susceptible to the 'dying ReLU' issue too. Shang *et al.* (2016) proposed a further slight improvement to the ReLU named 'Concatenated ReLU' (CReLU), allowing for both a positive and negative input activation, by applying ReLU after copying the input activations and concatenating them. Thus, CReLU is computationally expensive and prone to the 'dying ReLU' problem, although it generally leads to competitive classification performance with respect to the gold standard ReLU and LReLU AFs (Shang *et al.*, 2016).

**Table 1**. The main activation functions commonly used in deep neural networks, including the convolutional neural network (CNN), with their equations and reference. The ReLU and Leaky ReLU are the most common and reliable ones in CNNs.

| Activation function | Equation | Reference |
|---|---|---|
| Logistic Sigmoid | $f(x) = \dfrac{1}{1 + e^{-x}}$ | Han & Moraga (1995) |
| tanh | $f(x) = tanh(x) = \dfrac{2}{1 + e^{-2x}} - 1$ | Harrington (1993) |
| Softmax | $f_i(x) = \dfrac{e^{x_i}}{\sum_{j=1}^{J} e^{x_j}},$ $\forall\, i = [1, J]$ | Gold & Rangarajan (1996) |
| ArcTan | $f(x) = tan^{-1}(x)$ | Campbell *et al.* (1999) |
| SoftPlus | $f(x) = ln(1 + e^x)$ | Glorot *et al.* (2011) |
| Rectified Linear Unit (ReLU) | $f(x) = \begin{cases} 0\ \forall\, x < 0 \\ x\ \forall\, x \geq 0 \end{cases}$ | Nair & Hinton (2010) |
| Leaky Rectified Linear Unit (LReLU) | $f(x) = \begin{cases} x\ \forall\, x > 0 \\ \alpha x\ \forall\, x \leq 0, \\ \text{where}\ \alpha = 0.01 \end{cases}$ | Maas *et al.* (2013) |
| Parametric Rectified Linear Unit (PReLU) | $f(\alpha, x) = \begin{cases} \alpha x\ \forall\, x < 0 \\ x\ \forall\, x \geq 0 \end{cases}$ | He *et al.* (2015) |
| Exponential Linear Unit (ELU) | $f(\alpha, x) = \begin{cases} x\ \forall\, x > 0 \\ \alpha(e^x - 1)\ \forall\, x \leq 0 \end{cases}$ | Maas *et al.* (2013) |
| Scaled Exponential Linear Unit (SELU) | $f(\alpha, x) = \lambda \begin{cases} x\ \forall\, x \geq 0 \\ \alpha(e^x - 1)\ \forall\, x < 0, \\ \text{where}\ \lambda = 1.0507\ \text{and}\ \alpha = 1.67326 \end{cases}$ | Maas *et al.* (2013) |
| Concatenated Rectified Linear Unit (CReLU) | $f(x) = +\bigl(ReLU(x), ReLU(-x)\bigr)$ | Shang *et al.* (2016) |
| Flexible Rectified Linear Unit (FReLU) | $f(x) = \begin{cases} x + b_l\ \forall\, x > 0 \\ x\ \forall\, x \leq 0 \end{cases}$ | Qiu *et al.* (2018) |
| Randomized Rectified Linear Unit (RReLU) | $f(x) = y_{ji} = \begin{cases} x_{ji}\ \forall\, x_{ji} \geq 0 \\ a_{ji} x_{ji}\ \forall\, x_{ji} < 0 \end{cases}$ | Xu *et al.* (2015) |

## 1.3 Challenges with classical approaches to ReLU

Despite the wide application of DL-based algorithms for image classification in healthcare, such as the CNN (LeCun *et al.*, 2015) described in 1.2, its classical AF, although it mitigates the vanishing gradient issue typical of sigmoid AFs, can still experience the 'dying ReLU' problem. As discussed in 1.2, none of the recently proposed AFs, such as the LReLU, the PReLU, ELU and SeLU, have not solved this issue yet, as they are still algorithmically similar in their ReLU-like implementations.

This issue can lead to lack of generalisation for CNNs, thus hindering their application in a clinical setting. It is worth noting that, as an example, the last fully connected layer of the CNN in Kollias *et al.* (2018), having 1,500 neurons led, due to the 'dying ReLU' problem, to having only 30 neurons yielding non-zero values. Even by coupling a recurrent neural network (RNN) with their CNN, thus having a CNN-RNN (Kollias *et al.*, 2018), and their last layer then being designed with 128 neurons, only about 20 of them led to non-zero values, whilst the remaining ones experienced the 'dying ReLU' issue, yielding negligible values. These two examples confirm that classical approaches to ReLU failed to solve its associated 'dying ReLU' problem, thus warranting a different approach, which the authors suggest being of quantum nature, as illustrated in 1.4 and motivated in 1.5.

## 1.4 Quantum Machine Learning and related studies

Quantum ML is a relatively new field that blends the computational advantages brought by quantum computing and advances in ML beyond classical computation (Ciliberto, et al., 2018). Quantum ML has not only led to more effective algorithmic performance, but it has also enabled to find the global minimum in the solutions sought after in ML with a higher probability (Ciliberto, et al., 2018). The main principles of quantum computing are those inherited from quantum physics, such as superposition, entanglement, and interferences (Barabasi *et al.*, 2019). According to the quantum principle of superposition, the fundamental quantum bit or qubit can have multiple states at any point in time, i.e., a qubit can have a value of either 0 or 1, such as classical bits, but, differently from and beyond classical bits, a qubit can also have both values 0 and 1 concurrently (Barabasi *et al.*, 2019). A quantum gate is the unification of two quantum states for them to stay 'entangled' into an individual quantum state, wherein a change in one state would affect the other one and vice versa (Jozsa & Linden, 2003). Thus, a system of qubits, each of which holds multiple bits of information concurrently, behaves as one via the quantum property named 'entanglement', hence enabling massive parallelism too (Cleve *et al.*, 1998; Solenov *et al.*, 2018).

However, existing quantum approaches to implement AFs in deep neural networks have only adopted the repeat-until-success (RUS) technique to achieve pseudo non-linearity due to restrictions to linear and unitary operations in quantum mechanics (Nielsen & Chuang, 2002; Cao *et al.*, 2017). This RUS approach to AFs involves an individual state preparation routine and the generation of various superimposed and entangled linear combinations to propagate the routine of an AF to all states at unison. Thus, a deep neural network leveraging this quantum RUS technique could theoretically approximate most non-linear AFs (Macaluso *et al.*, 2020). Nevertheless, the practical applications of this approach are very limited due to the input range of the neurons in such architectures being bounded between 0 and $\pi/2$ as a trade-off of their theoretically generic AF formulation. Hu (2018) led a similar theoretical research effort in proposing a sigmoid-based non-linear AF, which is not periodic to enable a more efficient gradient descent whilst leveraging the principle of superposition in training neurons with multiple states concurrently. However, the classical form of the approach of Hu (2018) is the traditional ReLU, thus still not solving the 'dying ReLU' problem either. Konarac *et al.* (2020) leveraged a similar quantum-based sigmoid AF in their Quantum-Inspired Self-Supervised Network (QIS-Net) to provide high accuracy (99%) and sensitivity (96.1%) in magnetic resonance image segmentation, improving performance by about 1% with respect to classical approaches.

Differently from the related studies mentioned above, the two properties of entanglement and superposition could be pivotal in devising a quantum-based approach to ReLU in having both a positive solution and a negative one simultaneously, being able to avoid a negative solution by preferring the positive one, whereas traditional classical ReLU at times would fail by leading to negative solutions only, i.e., the 'dying ReLU' problem. Moreover, this principle enables quantum systems to reduce computational cost with respect to classical approaches, since several optimisations in multiple states can be performed concurrently (Schuld *et al.*, 2014).

## 1.5 Study rationale

As described in sections 1.1 and 1.2, DL is highly suitable in classifying medical images due to its intrinsic feature extraction mechanisms. As illustrated in both 1.2 and 1.4, the importance of the AF is evident in both classical and quantum DL, respectively. Although numerous variants of ReLU functions have been proposed in classical DL models (as revised in Section 1.2) they have not been widely adopted as ReLU and LReLU. These two AFs typically ensure accurate and reliable classification and are readily available in Python open source libraries, such as TensorFlow and Keras. Nevertheless, both these AFs and any recent AFs (see Section 1.2) have not solved the 'dying ReLU' problem yet. Moreover, vanishing and exploding gradient issues have not been fully resolved either. ELU and SELU may at times provide faster convergence than ReLU and LReLU, but they are not as reliable as those and are computationally more expensive (Pedamonti, 2018). Such unresolved issues lead to lack of generalisation that may hinder the diagnostic accuracy and reliability of an application leveraging DL techniques for the detection of COVID-19 or PD, thus resulting in a potentially high number of false negatives when the model's performance is evaluated on unseen patient data. The authors have hypothesised that this impaired generalisation is due to the classical approach underpinning such ReLU-based AFs that has been just leveraged and moulded in various ways so far, without breaking its inherent functional limitations.

The hereby contribution proposes, for the first time, that a quantum-based methodology to ReLU would improve the learning and generalisation in CNNs with relevant impact for critical applications, such as the above-mentioned diagnostic tasks. In particular, by blending the two key quantum principles of entanglement of qubits and the effects of superposition to help reach the global minimum in the solution, thus avoiding negative solutions differently from classical approaches as in 1.3, this study investigates the development of a novel AF 'Quantum ReLU' to avoid the problem of the 'dying ReLU' in a quantistic manner. This builds on recent research efforts by Cong *et al.* (2019) to develop a Quantum CNN that, although demonstrating how quantum states can be recognised, have not yet addressed the 'dying ReLU' problem, as it simply leveraged the traditional ReLUs instead.

Patterns from lung ultrasound images and spiral drawings are known diagnostic biomarkers for COVID-19 and PD respectively, PD being at times a delicate co-morbidity of COVID-19 patients, and improvements in generalisation are key to an accurate and reliable early diagnosis that can improve outcomes, especially in the event of co-morbidities. Thus, the novel Quantum ReLU will be leveraged in a CNN to improve classification performance in such pattern recognition tasks, as quantified via clinically relevant and interpretable metrics, and compared against the same CNN with current gold-standard AFs, including ReLU and LReLU. The proposed added capability of a Quantum ReLU in a CNN is hypothesised to improve its generalisation for pattern recognition in image classification, such as detecting COVID-19 and PD from ultrasound scans and spiral drawings, respectively.

The remaining sections of the paper are structured as follows. Section 2 deals with the methods, including sub-section 2.1 illustrating the two novel quantum AFs, along with their mathematical formulation and respective implementations in Python codes (in both TensorFlow and Keras libraries). Sub-section 2.2 provides a description of the benchmark datasets selected, along with a standardised data pre-processing strategy, whilst section 3 summarises the results obtained comparing the accuracy, reliability and computational time of a CNN with the proposed quantum AFs against salient gold standard AFs outlined in Table 1. Eventually, section 4 provides a thorough discussion of the results and section 5 summarises the current work and outlines its access, impact, and future applications.

## 2　METHODS

### 2.1　Two novel quantum activation functions

Despite appropriate initialisation of the weights to small random values via the He initialisation, with large weight updates, the summed input to the traditional ReLU activation function is always negative, although the input values fed to the CNN. Current improvements to the ReLU, such as the Leaky ReLU, allow for a more non-linear output to either account for small negative values or facilitate the transition from positive to small negative values, without eliminating the problem though.

Consequently, this study investigates the development of a novel activation function to obviate the problem of the 'dying ReLU' in a quantistic manner, i.e., by achieving a positive solution where previously the solution was negative. Such an added novel capability in a CNN was hypothesised to improve its generalisation for pattern recognition in image classification, particularly important in critical applications, such as medical diagnoses of COVID-19 and PD.

Thus, using the same standard two-layered CNN in TensorFlow for MNIST data classification, after identifying the main reproducible (with associated codes available in TensorFlow and Keras) AFs following a critical review of the literature (section 1), the following nine classical activation functions were considered: ReLU, Leaky ReLU, CReLU, sigmoid, tanh, softmax, VLReLU, ELU and SELU.

A two-step quantum approach was applied to ReLU first, by selecting its solution for positive values ($R(z) = z, \forall\, z > 0$), and the Leaky ReLU's solution for negative values ($R(z) = \alpha \times z, \forall\, z \leq 0, where\ \alpha = 0.01$) as a starting point to improve quantistically.

By applying the quantum principle of entanglement, the tensor product of the two candidate state spaces from ReLU and Leaky ReLU was performed and the following quantum-based combination of solutions was obtained:

$$R(z) = \alpha \times z - 2z, \forall\, z \leq 0 \tag{1}$$

Thus, keeping R(z) = z for positive values (z > 0) as in the ReLU, but with the added novelty of the entangled solution for negative values (1), the Quantum ReLU (QReLU) was attained (**Fig. 1**). The algorithms to describe the methodology and AF were implemented in TensorFlow and Keras, and presented in Listings 1 and 2 respectively, thus avoiding the 'dying ReLU' maintaining the positivity of the solution mathematically via this new quantum state.

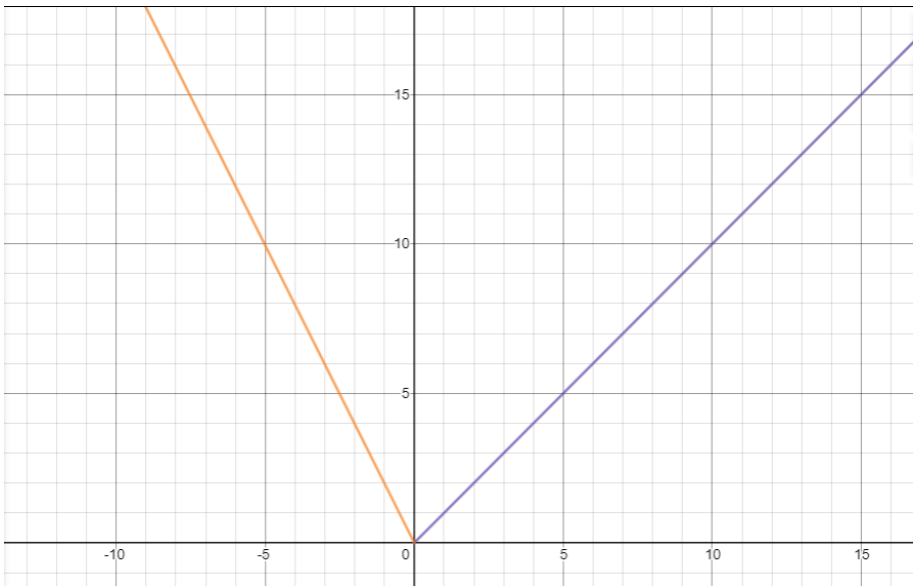

**Figure 1**. The Quantum ReLU (QReLU) activation function.

**Listing 1** provides the snippet of code in Python to leverage the QReLU in TensorFlow.

```python
# Defining the QReLU function
def q_relu(x):
  if x>0:
    x = x
    return x
  else:
    x = 0.01*x-2*x
    return x

# Vectorising the QReLU function
np_q_relu = np.vectorize(q_relu)

# Defining the derivative of the function QReLU
def d_q_relu(x):
  if x>0:
    x = 1
    return x
  else:
    x = 0.01-2
    return x

# Vectorising the derivative of the QReLU function
np_d_q_relu = np.vectorize(d_q_relu)

# Defining the gradient function of the QReLU
def q_relu_grad(op, grad):
    x = op.inputs[0]
    n_gr = tf_d_q_relu(x)
    return grad * n_gr

def py_func(func, inp, Tout, stateful=True, name=None, grad=None):
  # Generating a unique name to avoid duplicates:
    rnd_name = 'PyFuncGrad' + str(np.random.randint(0, 1E+2))
    tf.RegisterGradient(rnd_name)(grad)
```

```
    g = tf.get_default_graph()
    with g.gradient_override_map({"PyFunc": rnd_name}):
        return tf.py_func(func, inp, Tout, stateful=stateful, name=name)

np_q_relu_32 = lambda x: np_q_relu(x).astype(np.float32)
def tf_q_relu(x,name=None):
    with tf.name_scope(name, "q_relu", [x]) as name:
        y = py_func(np_q_relu_32,   # Forward pass function
                  [x],
                  [tf.float32],
                  name=name,
                   grad= q_relu_grad) # The function that overrides gradient
        y[0].set_shape(x.get_shape())     # To specify the rank of the input.
        return y[0]
np_d_q_relu_32 = lambda x: np_d_q_relu(x).astype(np.float32)
def tf_d_q_relu(x,name=None):
    with tf.name_scope(name, "d_q_relu", [x]) as name:
        y = tf.py_func(np_d_q_relu_32,
                  [x],
                  [tf.float32],
                  name=name,
                  stateful=False)
        return y[0]

# Example of usage in TensorFlow with a QReLU layer between a convolutional layer (#2) and a pooling layer (#2)
    conv2 = tf.layers.conv2d(
        inputs=pool1,
        filters=64,
        kernel_size=[5, 5],
        padding="same")
    conv2_act = tf_q_relu(conv2)
    pool2 = tf.layers.max_pooling2d(inputs=conv2_act, pool_size=[2, 2], strides=2)
```

**Listing 2** provides the snippet of code in Python to leverage the QReLU in Keras.

```python
# QReLU as a custom layer in Keras
from tensorflow.keras.layers import Layer

class QReLU(Layer):

    def __init__(self):
        super(QReLU,self).__init__()

    def build(self, input_shape):
        super().build(input_shape)

    def call(self, inputs,name=None):
        return tf_q_relu(inputs,name=None)

    def get_config(self):
        base_config = super(QReLU, self).get_config()
        return dict(list(base_config.items()))

    def compute_output_shape(self, input_shape):
        return input_shape

# Example of usage in a sequential model in Keras with a QReLU layer between a convolutional layer and a pooling layer

model = models.Sequential()
model.add(layers.Conv2D(32, (3, 3), input_shape=(32, 32, 3)))
model.add(QReLU())
model.add(layers.MaxPooling2D((2, 2)))
```

By leveraging the quantum principle of superposition on the QReLU's solution for positive and negative values, the following modified QReLU (m-QReLU) was obtained (**Fig. 2**). The algorithms to describe the methodology and AF were implemented in TensorFlow and Keras, and presented in Listings 3 and 4 respectively, still avoiding the 'dying ReLU' issue:

$$\alpha \times z - 2z + z = \alpha \times z - z, \forall z \leq 0 \qquad (2)$$

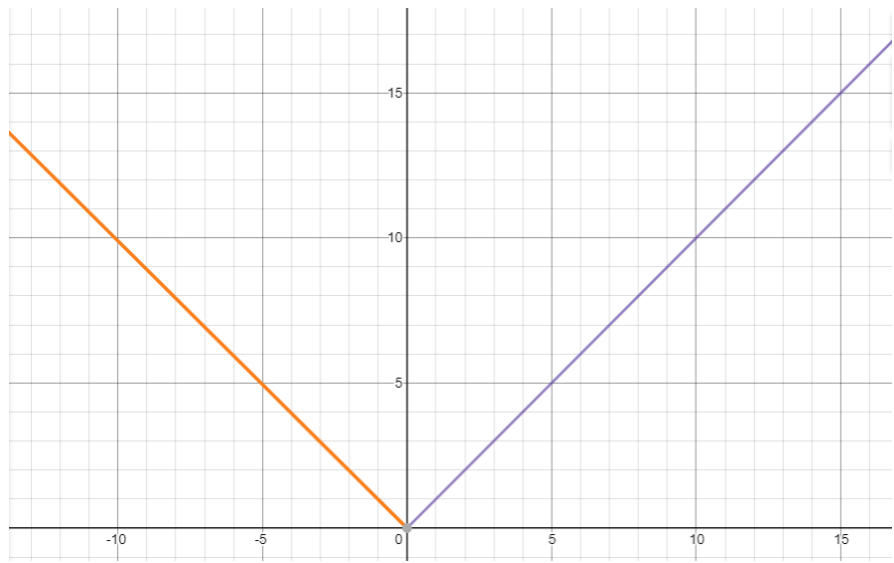

**Figure 2**. The modified Quantum ReLU (m-QReLU) activation function.

**Listing 3** provides the snippet of code in Python to leverage the m-QReLU in TensorFlow, using 'py_func' per **Listing 1**. Its usage in TensorFlow is the same as the 'QReLU' in Listing 1 but using 'tf_m_q_relu' as an activation function of the second convolutional layer ('conv2_act').

```
# Defining the m-QReLU (modified QReLU) function
def m_q_relu(x):
  if x>0:
    x = x
    return x
  else:
    x = 0.01*x-x
    return x

# Vectorising the m-QReLU function
np_m_q_relu = np.vectorize(m_q_relu)

# Defining the derivative of the function QReLU
def d_m_q_relu(x):
  if x>0:
    x = 1
    return x
  else:
    x = 0.01-1
    return x
```

```python
# Vectorising the derivative of the m-QReLU function
np_d_m_q_relu = np.vectorize(d_m_q_relu)

# Defining the gradient function of the QReLU
def m_q_relu_grad(op, grad):
    x = op.inputs[0]
    n_gr = tf_d_m_q_relu(x)
    return grad * n_gr

np_m_q_relu_32 = lambda x: np_m_q_relu(x).astype(np.float32)
def tf_m_q_relu(x,name=None):
    with tf.name_scope(name, "m_q_relu", [x]) as name:
        y = py_func(np_m_q_relu_32,   # Forward pass function
                [x],
                [tf.float32],
                name=name,
                 grad= m_q_relu_grad) # The function that overrides gradient
        y[0].set_shape(x.get_shape())     # To specify the rank of the input
        return y[0]
np_d_m_q_relu_32 = lambda x: np_d_m_q_relu(x).astype(np.float32)
def tf_d_m_q_relu(x,name=None):
    with tf.name_scope(name, "d_m_q_relu", [x]) as name:
        y = tf.py_func(np_d_m_q_relu_32,
                [x],
                [tf.float32],
                name=name,
                stateful=False)
        return y[0]
```

**Listing 4** provides the snippet of code in Python to leverage the m-QReLU in Keras.

```python
# QReLU as a custom layer in Keras
from tensorflow.keras.layers import Layer

class m_QReLU(Layer):

    def __init__(self):
        super(m_QReLU,self).__init__()

    def build(self, input_shape):
        super().build(input_shape)

    def call(self, inputs,name=None):
        return tf_m_q_relu(inputs,name=None)

    def get_config(self):
        base_config = super(m_QReLU, self).get_config()
        return dict(list(base_config.items()))

    def compute_output_shape(self, input_shape):
        return input_shape

# Example of usage in a sequential model in Keras with a m-QReLU layer between a convolutional layer and a pooling layer

model = models.Sequential()
model.add(layers.Conv2D(32, (3, 3), input_shape=(32, 32, 3)))
#model.add(QReLU())
model.add(m_QReLU())
model.add(layers.MaxPooling2D((2, 2)))
```

The m-QReLU also satisfies the entanglement principle being derived via the tensor outer product of the solutions from the QReLU.

Thus, a quantum-based blend of both superposition and entanglement principles mathematically leads the QReLU and the m-QReLU to obviate the 'dying ReLU' problem intrinsically. As shown in (1) and (2), although the two proposed AFs are quantistic in nature, both QReLU and m-QReLU can be run on classical hardware, such as central processing unit (CPU), graphics processing unit (GPU) and tensor processing unit (TPU), the latter being the type of runtime used in this study via Google Colab (http://colab.research.google.com/) to perform the required evaluation on the datasets described in 2.1. The novel QReLU and m-QReLU were developed and tested using Python 3.6 and written to be compatible with both TensorFlow (1.12 and 1.15 tested, 1.15 supports TensorFlow serving to deploy the novel AFs on the cloud) and the Keras Sequential API. Thus, both AFs were programmed as new Keras layers for ease of use.

By selecting the positive quantum state of the summed input of the QReLU and m-QReLU, an optimal early diagnosis could be achieved for patients with COVID-19 and PD. Thus, this study demonstrates the QReLU and m-QReLU as a potential new benchmark activation function to use in CNNs for critical image classification tasks, particularly useful in medical diagnoses, wherein generalisation is key to improving patient outcomes.

To assess which AF was suitable for each of the pattern recognition tasks involved in classifying the seven benchmark datasets as per 2.1, the performance of the baseline CNN was assessed via the test or out-of-sample classification accuracy, precision, sensitivity/recall and F1-score. Precision, recall, and F1-score are important metrics to measure the reliability of the classification outcomes. 95% confidence intervals (CIs) were also reported.

## 2.2 Benchmark data choices and pre-processing

To enable reproducibility and replicability of the results obtained, publicly available benchmark datasets were gathered and used in this study, as mentioned below. Moreover, to this purpose, full Python codes (.py and .ipynb formats) in both TensorFlow (https://www.tensorflow.org/) and Keras (https://keras.io/) on how these were used for training the model, as well as to evaluate its performance, are also provided.

As a general benchmark dataset for any image classifiers, especially CNNs, the MNIST data (LeCun *et al.*, 1998), including 60,000 images of handwritten digits (50,000 images for training, 10,000 images for testing), was used for the initial model and AF validation. This dataset is in tensor format available in TensorFlow (https://www.tensorflow.org/datasets/catalog/mnist).

To address the specific needs to improve diagnosis of Parkinson's disease (PD) and that of COVID-19 dealt with in this study, further benchmark datasets were used. Four benchmark datasets were leveraged to identify PD based on patterns on spiral drawings (1290 subjects in total), as follows:

- UCI Spiral Drawings dataset (data format: .jpg; Sakar *et al.*, 2013; Isenkul *et al.*, 2014) on 77 subjects (62 PD patients, 15 healthy controls) from the Department of Neurology in Cerrahpasa Faculty of Medicine, Istanbul University.
- Kaggle Spiral Drawings dataset (data format: .png; Zham *et al.*, 2017) on 55 subjects (27 PD patients, 28 healthy controls) from Dandenong Neurology, Melbourne, VIC, Australia.
- Spiral HandPD dataset (data format: .jpg; Pereira *et al.*, 2016a) on 92 subjects (74 PD patients, 18 healthy controls) from the Botucatu Medical School, São Paulo State University in Brazil.NewHandPD dataset (data format: .jpg; Pereira *et al.*, 2016b) on 66 additional subjects (31 PD patients, 35 healthy controls) – different from those in the 'Spiral HandPD dataset', from the Botucatu Medical School, São Paulo State University in Brazil.

Point-of-care lung ultrasound (POCUS) images (data format: .png) were considered to detect COVID-19; these include both benchmark datasets tested in this study (COVID-19 Ultrasound including US images from https://github.com/jannisborn/covid19_pocus_ultrasound/tree/master/data before July 22$^{nd}$, 2020, and POCUS 19 datasets having images from https://github.com/jannisborn/covid19_pocus_ultrasound/tree/master/data since July 22$^{nd}$, 2020) consisting in total of 1,108 images (659 on patients with COVID-19, 277 on patients with bacterial pneumonia and 175 on healthy lungs).

As in the MNIST dataset, images in all benchmark datasets were converted to grayscale and resized to be 28*28.

## 3   RESULTS

The two-layered CNN, designed as an MNIST classifier, was initially validated on the MNIST benchmark dataset itself, used for recognising handwritten digits. The QReLU and the m-QReLU were the best and second-best performing activation functions respectively, leading to an ACC and an F1-score of 0.99 (99%) and of 0.98 (98%) respectively (Table 2). The ReLU, the Leaky ReLU and the VLReLU also led to the best classification performance on the MNIST data (ACC = 0.99/99%, F1-score = 0.99/99%) (Table 2). Thus, the proposed QReLU achieved gold standard classification performance on this benchmark dataset.

Noteworthily, the QReLU and the m-QReLU led the same two-layered CNN architecture to achieve the best (ACC = 0.92/92%, F1-score = 0.93/93%) and third (ACC = 0.88/88%, F1-score = 0.90/90%) classification performance (Table 3) on the benchmark dataset named 'Spiral HandPD' on images of spiral drawings taken via graphic tablets from patients with PD and healthy subjects.

As illustrated in Table 4, competitive results were achieved by the QReLU and the m-QReLU versions on a further benchmark dataset on spiral drawings, the 'NewHandPD dataset', leading to the sixth and eight classification performance respectively (ACC = 0.83/83%, F1-score = 0.83/83%; ACC = 0.79/79%, F1-score = 0.79/79%). Very competitive outcomes were obtained by the two proposed quantum AFs on the Kaggle Spiral Drawings dataset, with m-QReLU (ACC = 0.73/73%, F1-score = 0.70/70%) and QReLU (ACC = 0.67/67%, F1-score = 0.67/67%) leading to the second and fourth classification performance respectively (Table 5), as well as when evaluated against the UCI Spiral Drawings dataset (QReLU ranked fifth with ACC = 0.82/82% and F1-score = 0.74/74%; m-QReLU ranked sixth with ACC = 0.78/78% and F1-score = 0.68/68%) (Table 6).

The overall increased generalisation brought about by the two novel quantum AFs is evident in the outstanding and mutually consistent classification outcomes achieved on both benchmark lung US datasets to distinguish COVID-19 from both pneumonia and healthy subjects with the best (Table 7 - QReLU and m-QReLU with ACC = 0.73/73% and F1-score = 0.73/73%) and the second (Table 8 - QReLU and m-QReLU with ACC = 0.6/60% and F1-score = 0.63/63%) classification performance respectively for both QReLU and m-QReLU.

Despite a higher computational cost (four-fold with respect to the other AFs except for the CReLU's increase being almost three-fold), the results achieved by either or both the proposed QReLU and m-ReLU AFs, assessed on classification accuracy, precision, recall and F1-score, indicate an overall higher generalisation achieved on five of the seven benchmark datasets (Table 2 on the MNIST data, Tables 3 and 5 on PD-related spiral drawings, Tables 7 and 8 on COVID-19 lung US images). Consequently, the two quantum ReLU methods are the overall best performing AFs that can be applied for aiding diagnosis of both COVID-19 from lung US data and PD from spiral drawings.

Specifically, when using the novel quantum AFs (QReLU and m-QReLU) as compared to the traditional ReLU and Leaky ReLU AFs, the gold standard AFs in DNNs, the following percentage increases in ACC, precision, recall/sensitivity and F1-score were noted:

- An increase of 55.32% in ACC and sensitivity/recall via m-QReLU as compared to ReLU and by 37.74% with respect to Leaky ReLU, thus avoiding the 'dying ReLU' problem when the CNN was evaluated on the Kaggle Spiral Drawings benchmark dataset (Table 5);
- An increase by 65.91% in F1-score via both QReLU and m-QReLU as opposed to Leaky ReLU, hence obviating the 'dying ReLU' problem again but when tested on the COVID-19 Ultrasound benchmark dataset (Table 7).
- An increase of 50% in ACC and sensitivity/recall via both QReLU and m-QReLU with regards to both ReLU and Leaky ReLU, hence solving the 'dying ReLU' problem when evaluated on the POCUS 19 Ultrasound benchmark dataset (Table 8).
- An increase by 82,000% in ACC and sensitivity/recall via QReLU (82%) when compared to tanh (0% ACC and sensitivity/recall), thus avoiding the vanishing gradient problem too, as assessed on the UCI Spiral Drawings benchmark dataset (Table 6).

Furthermore, it is worth noting the proposed quantum AFs led to improved classification outcomes as compared to recent advances in ReLU AFs, such as CReLU and VLReLU:

- QReLU led to ACC, precision, sensitivity/recall, and F1-score all higher by 1% those obtained via CReLU when evaluating the CNN's classification performance on the MNIST data (Table 2).
- m-QReLU resulted in an ACC and a sensitivity/recall higher by 3% than CReLU, and an F1-score greater by 2% on the Spiral HandPD dataset (Table 3).
- m-QReLU led to an ACC and a sensitivity/recall greater by 11% than VLReLU, and an F1-score also higher by 11% on the Spiral HandPD dataset (Table 3).
- m-QReLU resulted in an ACC and a sensitivity/recall higher by 6% than VLReLU, and an F1-score greater by 3% on the Kaggle Spiral Drawings dataset (Table 5).
- QReLU and m-QReLU led to an ACC and a sensitivity/recall greater by 9% and 18% than CReLU and VLReLU respectively, and an F1-score higher by 5% and 14% on the COVID-19 Ultrasound dataset (Table 7).
- QReLU and m-QReLU resulted in an ACC and a sensitivity/recall higher by 20% than VLReLU, and an F1-score greater by 10% on the POCUS 19 Ultrasound dataset (Table 8).

The results obtained via the QReLU and m-QReLU in a two-layered CNN on the MNIST dataset (Table 2) are comparable to those achieved on three- (LeCun *et al.*, 1998; Siddique *et al.*, 2019; Ahlawat *et al.*, 2020) and four-layered CNNs (Siddique *et al.*, 2019; Ahlawat *et al.*, 2020), as well as deeper architectures, such as ResNet and DenseNet (Chen *et al.*, 2018).

The two-layered CNN's classification performance via the proposed m-QReLU (ACC = 92%, F1-score = 93%, Table 3) was also higher by over 2% than the best performing five-layered CNNs, whose hyperparameters were also optimised respectively via both the Bat Algorithm and Particle Swarm Optimisation (PSO) (Pereira *et al.*, 2016c), to aid diagnosis of PD from spiral drawings, such as using the 'Spiral HandPD' benchmark dataset.

A comparable precision was achieved by the two-layered CNN model (Table 7) when the QReLU and m-QReLU were used as AFs with respect to the best classifier so far on the COVID 19 Ultrasound dataset, i.e., the sixteen-layered POCOVID-Net model, which builds on the VGG 16 model (Born *et al.*, 2020).

**Table 2**. Results on performance evaluation of the first Convolutional Neural Network having two convolutional layers, built in TensorFlow, and **tested on the MNIST benchmark dataset**.

| Activation function | Computational time* (s) | Accuracy (0-1) | Weighted average of precision (0-1) | Weighted average of recall (0-1) | Weighted average of F1-score (0-1) |
|---|---|---|---|---|---|
| ReLU | 725.79 | 0.99 (0.98-1.00) | 0.99 (0.99-1.00) | 0.99 (0.98-1.00) | 0.99 (0.98-0.99) |
| Leaky ReLU | 773.84 | 0.99 (0.98-1.00) | 0.99 (0.99-1.00) | 0.99 (0.98-1.00) | 0.99 (0.98-0.99) |
| CReLU | 1,214.08 | 0.98 (0.98-0.99) | 0.98 (0.98-0.99) | 0.98 (0.98-0.99) | 0.98 (0.98-0.99) |
| Sigmoid | 722.50 | 0.98 (0.95-0.99) | 0.98 (0.96-0.99) | 0.98 (0.96-0.99) | 0.98 (0.96-0.99) |
| Tanh | 725.51 | 0.98 (0.97-0.99) | 0.98 (0.98-0.99) | 0.98 (0.98-0.99) | 0.98 (0.98-0.99) |
| Softmax | 865.03 | 0.98 (0.97-0.99) | 0.98 (0.98-0.99) | 0.98 (0.98-0.99) | 0.98 (0.98-0.99) |
| VLReLU | 794.36 | 0.99 (0.98-1.00) | 0.99 (0.99-1.00) | 0.99 (0.98-1.00) | 0.99 (0.98-0.99) |
| ELU | 767.45 | 0.98 (0.97-0.99) | 0.98 (0.98-0.99) | 0.98 (0.98-0.99) | 0.98 (0.98-0.99) |
| SELU | 771.69 | 0.98 (0.98-0.99) | 0.98 (0.98-0.99) | 0.98 (0.98-0.99) | 0.98 (0.98-0.99) |
| **QReLU (This study)** | 3,369.73 | 0.99 (0.99-1.00) | 0.99 (0.99-1.00) | 0.99 (0.99-1.00) | 0.99 (0.99-1.00) |
| **m-QReLU (This study)** | 3,643.50 | 0.98 (0.98-0.99) | 0.98 (0.98-0.99) | 0.98 (0.98-0.99) | 0.98 (0.98-0.99) |

*It includes both training and evaluation.

CReLU: Concatenated ReLU; VLReLU: Very Leaky ReLU; ELU: Exponential Linear Unit; SELU: Scaled Exponential Linear Unit; QReLU: Quantum ReLU; m-QReLU: modified Quantum ReLU

**Table 3.** Results on performance evaluation of the first Convolutional Neural Network having two convolutional layers, built in Tensor-Flow, and **tested on the Spiral HandPD benchmark dataset**.

| Activation function | Computational time* (s) | Accuracy (0-1) | Weighted average of precision (0-1) | Weighted average of recall (0-1) | Weighted average of F1-score (0-1) |
|---|---|---|---|---|---|
| ReLU | 709.86 | 0.81 (0.76-0.87) | 1.00 (0.98-1.00) | 0.81 (0.78-0.83) | 0.90 (0.85-0.93) |
| Leaky ReLU | 783.76 | 0.82 (0.79-0.85) | 0.83 (0.81-0.85) | 0.82 (0.79-0.84) | 0.83 (0.80-0.84) |
| CReLU | 1,226.09 | 0.89 (0.87-0.92) | 0.95 (0.93-0.96) | 0.89 (0.87-0.91) | 0.91 (0.88-0.92) |
| Sigmoid | 717.86 | 0.85 (0.80-0.88) | 0.88 (0.85-0.90) | 0.85 (0.82-0.87) | 0.86 (0.83-0.88) |
| Tanh | 712.78 | 0.82 (0.81-0.85) | 0.92 (0.90-0.94) | 0.82 (0.79-0.84) | 0.86 (0.82-0.88) |
| Softmax | 799.64 | 0.85 (0.82-0.87) | 0.84 (0.82-0.86) | 0.85 (0.82-0.87) | 0.84 (0.83-0.86) |
| VLReLU | 771.20 | 0.81 (0.79-0.84) | 0.84 (0.82-0.86) | 0.81 (0.78-0.83) | 0.82 (0.80-0.84) |
| ELU | 743.69 | 0.86 (0.84-0.88) | 0.93 (0.91-0.94) | 0.86 (0.83-0.88) | 0.89 (0.86-0.91) |
| SELU | 759.33 | 0.85 (0.80-0.88) | 0.88 (0.85-0.90) | 0.85 (0.82-0.87) | 0.86 (0.83-0.88) |
| **QReLU (This study)** | 3,319.93 | 0.88 (0.87-0.90) | 0.95 (0.93-0.97) | 0.88 (0.87-0.90) | 0.90 (0.88-0.92) |
| **m-QReLU (This study)** | 3,393.56 | 0.92 (0.90-0.95) | 0.95 (0.94-0.96) | 0.92 (0.90-0.94) | 0.93 (0.91-0.95) |

*It includes both training and evaluation.

CReLU: Concatenated ReLU; VLReLU: Very Leaky ReLU; ELU: Exponential Linear Unit; SELU: Scaled Exponential Linear Unit; QReLU: Quantum ReLU; m-QReLU: modified Quantum ReLU

**Table 4**. Results on performance evaluation of the first Convolutional Neural Network having two convolutional layers, built in TensorFlow, and **tested on the NewHandPD benchmark dataset**.

| Activation function | Computational time* (s) | Accuracy (0-1) | Weighted average of precision (0-1) | Weighted average of recall (0-1) | Weighted average of F1-score (0-1) |
|---|---|---|---|---|---|
| ReLU | 540.81 | 0.75 (0.73-0.76) | 0.76 (0.74-0.78) | 0.75 (0.73-0.77) | 0.76 (0.74-0.77) |
| Leaky ReLU | 591.64 | 0.83 (0.82-0.84) | 0.83 (0.81-0.84) | 0.83 (0.81-0.84) | 0.83 (0.81-0.84) |
| CReLU | 928.91 | 0.92 (0.91-0.93) | 0.93 (0.91-0.94) | 0.92 (0.90-0.93) | 0.92 (0.91-0.93) |
| Sigmoid | 537.96 | 0.57 (0.53-0.59) | 1.00 (0.73-1.00) | 0.57 (0.54-0.59) | 0.72 (0.69-0.75) |
| Tanh | 541.33 | 0.94 (0.93-0.95) | 0.94 (0.92-0.95) | 0.94 (0.92-0.95) | 0.94 (0.92-0.95) |
| Softmax | 584.22 | 0.55 (0.51-0.57) | 1.00 (0.73-1.00) | 0.55 (0.52-0.57) | 0.71 (0.68-0.74) |
| VLReLU | 596.99 | 0.85 (0.83-0.86) | 0.87 (0.85-0.89) | 0.85 (0.83-0.87) | 0.85 (0.84-0.88) |
| ELU | 574.94 | 0.89 (0.88-0.90) | 0.89 (0.87-0.90) | 0.89 (0.87-0.91) | 0.89 (0.87-0.90) |
| SELU | 571.76 | 0.85 (0.83-0.86) | 0.86 (0.84-0.88) | 0.85 (0.83-0.87) | 0.85 (0.84-0.87) |
| **QReLU** (This study) | 3,407.02 | 0.83 (0.82-0.85) | 0.83 (0.81-0.86) | 0.83 (0.82-0.84) | 0.83 (0.82-0.85) |
| **m-QReLU** (This study) | 3,453.62 | 0.79 (0.78-0.81) | 0.79 (0.77-0.82) | 0.79 (0.78-0.80) | 0.79 (0.78-0.81) |

*It includes both training and evaluation.

CReLU: Concatenated ReLU; VLReLU: Very Leaky ReLU; ELU: Exponential Linear Unit; SELU: Scaled Exponential Linear Unit; QReLU: Quantum ReLU; m-QReLU: modified Quantum ReLU

**Table 5**. Results on performance evaluation of the first Convolutional Neural Network having two convolutional layers, built in TensorFlow, and **tested on the Kaggle Spiral Drawings benchmark dataset**. The size of the images was set to 28*28, as per the MNIST benchmark dataset.

| Activation function | Computational time* (s) | Accuracy (0-1) | Weighted average of precision (0-1) | Weighted average of recall (0-1) | Weighted average of F1-score (0-1) |
|---|---|---|---|---|---|
| ReLU | 700.16 | 0.47 (0.45-0.49) | 0.47 (0.45-0.49) | 0.47 (0.45-0.49) | 0.47 (0.45-0.49) |
| Leaky ReLU | 762.75 | 0.53 (0.51-0.55) | 0.54 (0.52-0.56) | 0.53 (0.51-0.55) | 0.54 (0.52-0.55) |
| CReLU | 1,203.73 | 0.80 (0.79-0.83) | 0.84 (0.82-0.85) | 0.80 (0.78-0.83) | 0.80 (0.79-0.83) |
| Sigmoid | 707.94 | 0.50 (0.46-0.52) | 1.00 (0.69-1.00) | 0.50 (0.47-0.52) | 0.67 (0.64-0.70) |
| Tanh | 700.40 | 0.63 (0.61-0.65) | 0.65 (0.63-0.68) | 0.63 (0.60-0.65) | 0.64 (0.61-0.66) |
| Softmax | 790.07 | 0.63 (0.62-0.65) | 0.64 (0.62-0.66) | 0.63 (0.60-0.65) | 0.63 (0.61-0.65) |
| VLReLU | 759.44 | 0.67 (0.65-0.69) | 0.67 (0.65-0.69) | 0.67 (0.65-0.69) | 0.67 (0.65-0.69) |
| ELU | 734.69 | 0.50 (0.46-0.52) | 0.50 (0.46-0.52) | 0.50 (0.46-0.52) | 0.50 (0.46-0.52) |
| SELU | 759.05 | 0.67 (0.66-0.70) | 0.70 (0.68-0.71) | 0.67 (0.65-0.69) | 0.67 (0.66-0.70) |
| **QReLU (This study)** | 3,527.64 | 0.67 (0.66-0.69) | 0.68 (0.66-0.70) | 0.67 (0.65-0.69) | 0.67 (0.66-0.69) |
| **m-QReLU (This study)** | 3,540.10 | 0.73 (0.72-0.76) | 0.74 (0.69-0.76) | 0.73 (0.68-0.75) | 0.70 (0.69-0.73) |

*It includes both training and evaluation.

CReLU: Concatenated ReLU; VLReLU: Very Leaky ReLU; ELU: Exponential Linear Unit; SELU: Scaled Exponential Linear Unit; QReLU: Quantum ReLU; m-QReLU: modified Quantum ReLU

**Table 6**. Results on performance evaluation of the first Convolutional Neural Network having two convolutional layers, built in TensorFlow, and **tested on the University California Irvine (UCI) Spiral Drawings benchmark dataset**. The Kaggle Spiral Drawings benchmark dataset, which includes drawings from both healthy subjects and patients with Parkinson's Disease, was used for training and the UCI Spiral Drawings benchmark dataset, which only has spiral drawings acquired during both static and dynamic tests from patients with PD, was deployed for testing. The size of the images was set to 28*28, as per the MNIST benchmark dataset.

| Activation function | Computational time* (s) | Accuracy (0-1) | Weighted average of precision (0-1) | Weighted average of recall (0-1) | Weighted average of F1-score (0-1) |
|---|---|---|---|---|---|
| ReLU | 536.19 | 0.78 (0.74-0.81) | 0.61 (0.59-0.64) | 0.78 (0.75-0.81) | 0.68 (0.64-0.72) |
| Leaky ReLU | 595.66 | 0.76 (0.72-0.78) | 0.58 (0.55-0.62) | 0.76 (0.73-0.80) | 0.66 (0.64-0.70) |
| CReLU | 930.60 | 0.96 (0.95-0.97) | 0.92 (0.91-0.94) | 0.96 (0.95-0.97) | 0.94 (0.92-0.95) |
| Sigmoid | 543.40 | 1.00 (0.99-1.00) | 1.00 (0.99-1.00) | 1.00 (0.99-1.00) | 1.00 (0.99-1.00) |
| Tanh | 528.71 | 0.00 (0.00-0.02) | 0.00 (0.00-0.02) | 0.00 (0.00-0.03) | 0.00 (0.00-0.02) |
| Softmax | 585.93 | 1.00 (0.99-1.00) | 1.00 (0.99-1.00) | 1.00 (0.99-1.00) | 1.00 (0.99-1.00) |
| VLReLU | 751.23 | 0.86 (0.83-0.88) | 0.74 (0.72-0.76) | 0.86 (0.84-0.88) | 0.80 (0.74-0.82) |
| ELU | 734.46 | 0.52 (0.36-0.54) | 0.27 (0.24-0.29) | 0.52 (0.35-0.54) | 0.36 (0.26-0.43) |
| SELU | 730.00 | 0.62 (0.45-0.64) | 0.38 (0.34-0.42) | 0.62 (0.46-0.64) | 0.47 (0.39-0.56) |
| **QReLU (This study)** | 3,449.55 | 0.82 (0.79-0.84) | 0.67 (0.65-0.70) | 0.82 (0.78-0.84) | 0.74 (0.68-0.77) |
| **m-QReLU (This study)** | 3,325.67 | 0.78 (0.75-0.81) | 0.61 (0.60-0.65) | 0.78 (0.76-0.81) | 0.68 (0.66-0.74) |

*It includes both training and evaluation.

CReLU: Concatenated ReLU; VLReLU: Very Leaky ReLU; ELU: Exponential Linear Unit; SELU: Scaled Exponential Linear Unit; QReLU: Quantum ReLU; m-QReLU: modified Quantum ReLU

**Table 7**. Results on performance evaluation of the first Convolutional Neural Network having two convolutional layers, built in TensorFlow, and **tested on the COVID-19 Ultrasound benchmark dataset**. The size of the images was set to 28*28, as per the MNIST benchmark dataset.

| Activation function | Computational time* (s) | Accuracy (0-1) | Weighted average of precision (0-1) | Weighted average of recall (0-1) | Weighted average of F1-score (0-1) |
|---|---|---|---|---|---|
| ReLU | 739.97 | 0.64 (0.62-0.67) | 0.79 (0.77-0.82) | 0.64 (0.62-0.66) | 0.70 (0.64-0.73) |
| Leaky ReLU | 799.42 | 0.45 (0.42-0.48) | 0.45 (0.42-0.47) | 0.45 (0.43-0.46) | 0.44 (0.42-0.46) |
| CReLU | 1,290.60 | 0.64 (0.62-0.67) | 0.73 (0.70-0.76) | 0.64 (0.62-0.66) | 0.68 (0.65-0.71) |
| Sigmoid | 737.86 | 0.64 (0.62-0.67) | 0.73 (0.70-0.76) | 0.64 (0.62-0.66) | 0.66 (0.64-0.69) |
| Tanh | 736.93 | 0.55 (0.52-0.58) | 0.67 (0.63-0.69) | 0.55 (0.53-0.58) | 0.59 (0.56-0.62) |
| Softmax | 841.28 | 0.45 (0.42-0.49) | 0.61 (0.57-0.65) | 0.45 (0.43-0.47) | 0.51 (0.47-0.56) |
| VLReLU | 780.88 | 0.55 (0.52-0.58) | 0.67 (0.63-0.69) | 0.55 (0.53-0.58) | 0.59 (0.56-0.62) |
| ELU | 760.62 | 0.64 (0.62-0.67) | 0.61 (0.58-0.65) | 0.64 (0.62-0.66) | 0.60 (0.57-0.63) |
| SELU | 758.67 | 0.64 (0.62-0.67) | 0.73 (0.70-0.76) | 0.64 (0.62-0.66) | 0.65 (0.63-0.68) |
| **QReLU (This study)** | 3,429.00 | 0.73 (0.72-0.76) | 0.76 (0.74-0.78) | 0.73 (0.71-0.75) | 0.73 (0.72-0.76) |
| **m-QReLU (This study)** | 3,465.98 | 0.73 (0.72-0.76) | 0.76 (0.74-0.78) | 0.73 (0.71-0.75) | 0.73 (0.72-0.76) |

*It includes both training and evaluation.

CReLU: Concatenated ReLU; VLReLU: Very Leaky ReLU; ELU: Exponential Linear Unit; SELU: Scaled Exponential Linear Unit; QReLU: Quantum ReLU; m-QReLU: modified Quantum ReLU

**Table 8**. Results on performance evaluation of the first Convolutional Neural Network having two convolutional layers, built in TensorFlow, trained on the COVID-19 Ultrasound benchmark dataset (from Table 4) and **tested on the POCUS 19 benchmark dataset**.

| Activation function | Computational time* (s) | Accuracy (0-1) | Weighted average of precision (0-1) | Weighted average of recall (0-1) | Weighted average of F1-score (0-1) |
|---|---|---|---|---|---|
| ReLU | 720.63 | 0.40 (0.33-0.47) | 1.00 (0.74-1.00) | 0.40 (0.35-0.44) | 0.57 (0.52-0.61) |
| Leaky ReLU | 781.64 | 0.40 (0.32-0.46) | 0.80 (0.77-0.84) | 0.40 (0.35-0.44) | 0.53 (0.47-0.57) |
| CReLU | 1,229.20 | 0.60 (0.59-0.62) | 0.87 (0.85-0.89) | 0.60 (0.58-0.63) | 0.63 (0.61-0.67) |
| Sigmoid | 733.05 | 0.40 (0.33-0.47) | 1.00 (0.74-1.00) | 0.40 (0.35-0.44) | 0.57 (0.52-0.61) |
| Tanh | 730.13 | 0.40 (0.33-0.47) | 1.00 (0.74-1.00) | 0.40 (0.35-0.44) | 0.57 (0.52-0.61) |
| Softmax | 825.47 | 0.60 (0.59-0.62) | 0.87 (0.85-0.89) | 0.60 (0.58-0.63) | 0.63 (0.61-0.67) |
| VLReLU | 781.89 | 0.40 (0.32-0.46) | 0.80 (0.77-0.84) | 0.40 (0.35-0.44) | 0.53 (0.47-0.57) |
| ELU | 765.93 | 0.40 (0.33-0.47) | 1.00 (0.74-1.00) | 0.40 (0.35-0.44) | 0.57 (0.52-0.61) |
| SELU | 766.93 | 0.80 (0.77-0.84) | 0.87 (0.85-0.89) | 0.80 (0.78-0.84) | 0.80 (0.76-0.83) |
| **QReLU (This study)** | 3,407.86 | 0.60 (0.59-0.62) | 0.87 (0.85-0.89) | 0.60 (0.58-0.63) | 0.63 (0.61-0.67) |
| **m-QReLU (This study)** | 3,196.74 | 0.60 (0.59-0.62) | 0.87 (0.85-0.89) | 0.60 (0.58-0.63) | 0.63 (0.61-0.67) |

*It includes both training and evaluation.

CReLU: Concatenated ReLU; VLReLU: Very Leaky ReLU; ELU: Exponential Linear Unit; SELU: Scaled Exponential Linear Unit; QReLU: Quantum ReLU; m-QReLU: modified Quantum ReLU

## 4 DISCUSSION

Further to the extensive review of existing ReLU AFs provided in Section 1.2, also considering that classical approaches have been unable to solve the 'dying ReLU' problem as reviewed in Section 1.3, and taking into account the advantages of quantum states in AFs (listed in Section 1.4), two novel quantum-based AFs were mathematically formulated in Section 2.2 and developed in both TensorFlow (**Listings 1 and 3**, https://www.tensorflow.org/) and Keras (**Listings 2 and 4**, https://keras.io/) to enable reproducibility and replicability. Thus, the MNIST two-layered CNN-based classifier in TensorFlow was selected as the baseline model to assess the impact of using either quantum AFs (QReLU and m-QReLU) on the classification performance on seven benchmark datasets as described in Section 2.1 and evaluated based on test ACC, precision, recall/sensitivity and F1-score, as mentioned in Section 2.2.

The proposed QReLU leads to the best classification performance on the MNIST benchmark dataset (ACC = 99%, F1-score = 99%, Table 2) to recognise handwritten digits serves as a regression test to validate the hypothesis whereby, using the baseline CNN-based MNIST classifier, the highest classification performance is achieved with the presumed best AF. This hypothesis has been further confirmed by the m-QReLU achieving the second classification performance (ACC = 99%, F1-score = 99%, Table 2) across all eleven AFs evaluated as in 2.2. Achieving the same classification performance as the gold standard reproducible and replicable AFs in CNNs (ReLU, the Leaky ReLU and the VLReLU) – readily available in both TensorFlow and Keras – the QReLU can be granted the designation of benchmark AF for the task of handwritten digits recognition performed on the MNIST benchmark dataset.

The benefits of avoiding the 'dying ReLU' problem become evident when assessing the same two-layered CNN architecture with the QReLU especially (ACC = 0.92/92%, F1-score = 0.93/93%, Table 3), which achieved the best classification performance on critical image classification tasks, such as recognising PD-related patterns from spiral drawings in the 'Spiral HandPD' benchmark dataset. The higher generalisability achieved via the two proposed quantum AFs in further support of the advantage of obviating the 'dying ReLU' issue is evident from the best classification performance in differentiating COVID-19 from both bacterial pneumonia and healthy controls from the Lung US data (Table 7 - QReLU and m-QReLU with ACC = 0.73/73% and F1-score = 0.73/73%). Such an overall higher diagnostic performance is corroborated by the second-best classification outcomes attained on the second benchmark Lung US dataset (Table 8).

Whilst traditional ReLU approaches show highly variable classification outcomes, especially when they experience the 'dying ReLU' problem (Tables 5, 7 and 8), both the QReLU and the m-QReLU were able to ensure a consistently higher classification performance and generalisation across the entire variety of image classification tasks involved, from the benchmark handwritten digits recognition task (MNIST), to recognising PD-related patterns from spiral drawings taken from graphic tablets, to aiding detection of COVID-19 from bacteria pneumonia and healthy lungs based on US scans. The advantage of using the proposed AFs for COVID-19 detection lies in the potential for their translational applications in a clinical setting, i.e., in leveraging CNNs with the QReLU or m-QReLU to detect COVID-19 in patients with neurodegenerative co-morbidities, such as PD, via non-ionising medical imaging (e.g., US). This added capability will come handy in future, as portable MRI and ML-enhanced MRI technologies will also become more affordable and widespread, thus being improvable with deep learning models (e.g., the two-layered CNN with QReLU or m-QReLU AFs in this study). Solutions either on edge devices or on the cloud for tele-diagnosis and tele-monitoring required in pandemics similar to the current one (COVID-19) could be soon suitable for in-home diagnostic and prognostic assessments too, which should improve personalised care for shielded or vulnerable individuals.

Moreover, competitive outcomes were obtained via the QReLU and the m-QReLU on three further benchmark datasets, e.g., 'NewHandPD dataset', the Kaggle and the UCI Spiral Drawings benchmark datasets, with ACC and F1-score mostly above 75% (Tables 4-6) using the relatively simple deep neural network leveraged in this study (the two-layered MNIST CNN classifier). Such results also demonstrate the added capability of the proposed QReLU and the m-QReLU to avoid the vanishing gradient problem occurred using tanh (0% ACC and sensitivity/recall), as evaluated on the UCI Spiral Drawings benchmark dataset (Table 6).

Despite the overall increase in generalisability brought about by the QReLU and the m-QReLU, the computational cost of the CNN increased by four times as compared to the other nine AFs evaluated, except for the CReLU, against which a three-fold increase was reported (Tables 2-8). Nevertheless, considering the importance of achieving higher classification performance over lower computational cost for diagnostic applications in a clinical setting, especially for the critical image classification tasks involved in this study, such as the detection of PD (Tables 3-6) and COVID-19 (Tables 7 and 8), this increase in computational cost is not expected to impair the wide application of the two novel quantum AFs to aid such diagnostic tasks and any other medical applications involving image classification.

In fact, the QReLU and m-QReLU have been demonstrated as considerably better than the current (undisputedly assumed) gold standard AFs in CNNs, i.e., the traditional ReLU and the Leaky ReLU. In particular, an increase by 50-66% in both accuracy and reliability (especially, sensitivity/recall and F1-score) metrics was reported across both pattern recognition tasks, i.e., detection of PD-related patterns from spiral drawings (Tables 5 and 6) and aiding diagnosis of COVID-19 from US scans (Table 7). The two proposed quantum AFs also outperformed more cutting-edge ReLU AFs, such as the CReLU and the VLReLU, by 5-20% across all classification tasks considered, i.e., MNIST data classification (Table 2), spiral drawings PD-related pattern recognition (in particular, Tables 3 and 5), and COVID-19 detection from US scans (Tables 7 and 8).

Moreover, the QReLU and the m-QReLU led the baseline two-layered CNN MNIST classifier to achieve a comparable classification performance on the MNIST dataset as deeper CNNs, ranging from three to four layers (LeCun *et al.*, 1998; Siddique *et al.*, 2019; Ahlawat *et al.*, 2020), including deeper architectures, e.g., ResNet and DenseNet (Chen *et al.*, 2018). It is worth noting that, when leveraging the QReLU and the m-QReLU, the two-layered CNN with hyperparameters based on the MNIST data outperformed (ACC = 92%, F1-score = 93%, Table 3) deeper and BA- and PSO-optimised CNNs from published studies by over 2% (Pereira *et al.*, 2016c) in aiding the diagnosis of PD from patterns in spiral drawings (e.g., using the 'Spiral HandPD' benchmark data). The two-layered CNN model with either QReLU or m-QReLU as AFs achieved a comparable precision (Table 7) to the best-performing classifier on the COVID 19 Ultrasound dataset, i.e., the sixteen-layered POCOVID-Net model, which is an extension of the VGG 16 benchmark model (Born *et al.*, 2020).

These outcomes show the two main practical advantages brought about by the avoidance of the 'dying ReLU' problem in QReLU and the m-QReLU that outweigh the initial consideration on these two quantum AFs leading to an overall higher computational cost despite the increased generalisation, which are as follows:

1. Using QReLU or m-QReLU can obviate the need for several convolutional layers in CNNs and any CNN-derived models, such as AlexNet, ResNet, DenseNet, CondenseNet, cCondensenet and VGG 16, as demonstrated above and in section 3 (results),
2. Leveraging QReLU or m-QReLU as AFs in CNN can minimise the need for optimisation of CNN's hyperparameters.

The implications of the two above-mentioned practical benefits are multiple. Firstly, the two proposed AFs may not only improve generalisation but also computational cost when considering image classification tasks that involve deeper architectures than the two-layered CNN used in this study. Thus, the proposed AFs may be viable alternatives to the ReLU AF, which is the current gold standard AF in CNNs. Second, by improving both generalisation and computational cost when deeper architectures may be required, the QReLU and m-QReLU may be suitable for tasks that require scalability of deep neural networks. Third, the proposed quantum AFs may enable more effective transfer learning, such as for COVID-19 detection in multiple geographical areas, as well as extending trained deep nets to further diagnostic tasks, including prognostic applications too, and aiding self-driving vehicles in image classification tasks essential to ensure passenger safety.

Overall, the avoidance of the 'dying ReLU' problem achieved via QReLU and m-QReLU is expected to radically shift the paradigm of blindly relying on the traditional ReLU AF in CNN and any CNN-derived models, and embrace innovative approaches, including quantum-based, such as the two novel AFs designed, developed and validated in this study.

## 5   Conclusion

Further to a thorough analysis of the classification performance of the two-layered CNN MNIST classifier leveraging the two quantum AFs developed in this study, QReLU and m-QReLU, and evaluated against nine benchmark AFs, including ReLU and its main recent reproducible and replicable advances, as well as relevant published studies, the proposed QReLU and m-QReLU prove to be the first two AFs in the recorded history of deep learning to successfully avoid the 'dying ReLU' problem, by design. Their novel algorithms describing the methodology and AF were implemented in TensorFlow and Keras, as well as presented in Listings 1-4. This added capability ensured accurate and reliable classification for recognising PD-related patterns from spiral drawings and detecting COVID-19 from non-ionising medical imaging (US) data.

Furthermore, its availability in both Google's TensorFlow and Keras – the two most popular libraries in Python for deep learning - facilitate their wide application beyond clinical diagnostics, including medical prognostics and any other applications involving image classification. Thus, the QReLU and m-QReLU can aid detection of COVID-19 during these unprecedented times of this pandemic, as well as deliver continuous value added in aiding the diagnosis of PD based on pattern recognition from spiral drawings.

Noteworthily, when leveraging the proposed quantum AFs, the baseline CNN model achieved comparable classification performance to deeper CNN and CNN-derived architectures across all image recognition tasks involved in this study, from handwritten digits recognition, to detection of PD-related patterns from spiral drawings and COVID-19 from lung US scans. Thus, these outcomes corroborate the benefit of using AFs that avoid the 'dying ReLU' problem for critical image classification tasks, such as for medical diagnoses, making them a viable alternative to the current gold standard AF in CNNs, i.e., the ReLU. This study is expected to have a radical impact in redefining the benchmark AFs in CNN and CNN-derived deep learning architectures for applications across academic research and industry.

**Acknowledgment**
The authors would like to thank two research assistants from the University of Bradford, Ms Smriti Kotiyal and Mr Rohit Trivedi, for their assistance to the background review relevant for this paper.

The authors declare that no ethical approval was required for carrying out the study, as the data used in it were taken from publicly available repositories and appropriately referenced in text. Moreover, the authors declare not to have any competing interests and an appropriate funding statement has been provided on the title page of this article.

**REFERENCES**

Ahlawat, S., Choudhary, A., Nayyar, A., Singh, S., & Yoon, B. (2020). Improved Handwritten Digit Recognition Using Convolutional Neural Networks (CNN). *Sensors*, 20(12), 3344.

Barabasi, I., Tappert, C. C., Evans, D., & Leider, A. M. (2019). Quantum Computing and Deep Learning Working Together to Solve Optimization Problems. 2019 *International Conference on Computational Science and Computational Intelligence* (*CSCI*).

Beam, A. L., & Kohane, I. S. (2018). Big Data and Machine Learning in Health Care. *JAMA*, 319(13), 1317-1318.

Beauchamp, L. C., Finkelstein, D. I., Bush, A. I., Evans, A. H., & Barnham, K. J. (2020). Parkinsonism as a Third Wave of the COVID-19 Pandemic?. *Journal of Parkinson's Disease*, (Preprint), 1-11.

Bhaskar, S., Bradley, S., Israeli-Korn, S., B. M., Chattu, V. K., Thomas, P., Mart, S. (2020). Chronic Neurology in COVID-19 Era: Clinical Considerations and Recommendations from the REPROGRAM Consortium. *Front. Neurol*.

Born, J., Brändle, G., Cossio, M., Disdier, M., Goulet, J., Roulin, J., & Wiedemann, N. (2020). POCOVID-Net: automatic detection of COVID-19 from a new lung ultrasound imaging dataset (POCUS). *arXiv*:2004.12084.

Campbell, S. A., Ruan, S., Wolkowicz, G., & Wu, J. (1999). Stability and bifurcation of a simple neural network with multiple time delays. *Fields Inst. Commun*, 21(4), 65-79.

Cao, Y., Guerreschi, G. G., & Aspuru-Guzik, A. (2017). Quantum Neuron: an elementary building block for machine learning on quantum computers. *arXiv*.

Centers for Disease Control and Prevention. (2020, May 13). Retrieved from *Coronavirus disease*: https://www.cdc.gov/coronavirus/2019-ncov/symptoms-testing/symptoms.html.

Chen, F., Chen, N., Mao, H., & Hu, H. (2018). Assessing four neural networks on handwritten digit recognition dataset (MNIST). *arXiv*:1811.08278.

Ciliberto, C., Herbster, M., Ialongo, A. D., Pontil, M., Rocchetto, A., Severini, S., Wossnig, L. (2018). Quantum machine learning: a classical perspective. *Proc. Royal Soc*, 474: 20170551.

Cleve, R., Ekert, A., Macchiavello, C., & Mosca, M. (1998). Quantum algorithms revisited. *The Royal Society*, 454, 339–354.

Cong, I., Choi, S., & Lukin, M. D. (2019). Quantum Convolutional Neural Networks. *Nature Physics*, 15, 1273–1278.

Cui, N., 2018. Applying Gradient Descent in Convolutional Neural Networks. *Journal of Physics: Conference Series*, 1004, p.012027.

Cui, J., Li, F., & Shi, Z.-L. (2019). Origin and evolution of pathogenic coronaviruses. *Nature Reviews Microbiology*, 17, 181-192.

Gao, H., Cai, L., & Ji, S. (2020). Adaptive Convolution ReLUs. *Thirty-Fourth AAAI Conference on Artificial Intelligence*.

Glorot, X., Bordes, A., & Bengio, Y. (2011, June). Deep sparse rectifier neural networks. In *Proceedings of the fourteenth international conference on artificial intelligence and statistics*, 315-323.

Gold, S., & Rangarajan, A. (1996). Softmax to softassign: Neural network algorithms for combinatorial optimization. *Journal of Artificial Neural Networks*, 2(4), 381-399.

Han, J., & Moraga, C. (1995, June). The influence of the sigmoid function parameters on the speed of backpropagation learning. In *International Workshop on Artificial Neural Networks*, 195-201. Springer, Berlin, Heidelberg.

Harrington, P. D. B. (1993). Sigmoid transfer functions in backpropagation neural networks. *Analytical Chemistry*, 65(15), 2167-2168.

He, K., Zhang, X., Ren, S., & Sun, J. (2015). Delving Deep into Rectifiers: Surpassing Human-Level Performance on ImageNet Classification. *IEEE International Conference on Computer Vision (ICCV)*. Santiago.

Hinton, G. E., & Salakhutdinov, R. R. (2006). Reducing the Dimensionality of Data with Neural Networks. *Science*, 504-507, 313.

Hu, W. (2018). Towards a Real Quantum Neuron. *Natural Science*, 10 (3).

Isenkul, M., Sakar, B., & Kursun, O. (2014, May). Improved spiral test using digitized graphics tablet for monitoring Parkinson's disease. In *Proc. of the Int'l Conf. on e-Health and Telemedicine*, 171-175.

Jozsa, R., & Linden, N. (2003). On the role of entanglement in quantum-computational speed-up. *The Royal Society*, 10.1098.

Ker, J., Wang, L., Rao, J., & Lim, T. (2017). Deep Learning Applications in Medical Image Analysis. *IEEE Access*, 6: 9375 - 9389.

Klambauer, G., Unterthiner, T., & Mayr, A. (2017). Self-Normalizing Neural Networks. *arXiv*.

Kollias, D., Tagaris, A., Stafylopatis, A., Kollias, S., & Tagaris, G. (2018). Deep neural architectures for prediction in healthcare. *Complex & Intelligent Systems*, 4, 119-131.

Konarac, D., Bhattacharyya, S., Gandhia, T. K., & Panigrahia, B. K. (2020). A Quantum-Inspired Self-Supervised Network model for automatic segmentation of brain MR images. *Applied Soft Computing*, 93, 106348.

Krizhevsky, A., Sutskever, I., & Hinton, G. E. (2012). ImageNet Classification with Deep Convolutional Neural Networks. *Proceedings of the Advances in Neural Information Processing Systems*. 1097-1105.

LeCun, Y. & Bengio, Y. (1995). Convolutional networks for images, speech, and time-series. In M. A. Arbib (Ed.), The handbook of brain theory and neural networks MIT Press.

LeCun, Y., Bottou, L., Bengio, Y., & Haffner, P. (1998). Gradient-based learning applied to document recognition. *Proceedings of the IEEE*, 86(11), 2278-2324.

LeCun, Y., Bengio, Y. & Hinton, G. (2015). Deep Learning. *Nature*. 521. 436-44.

Liew, S. S., Khalil-Hani, M., & Bakhteri, R. (2016). Bounded activation functions for enhanced training stability of deep neural networks on visual pattern recognition problems. *Neurocomputing*, 216, 718-734.

Litjens, G., Kooi, T., Bejnordi, B. E., Setio, A. A., Ciompi, F., Ghafoorian, M., Sánchez, C. I. (2017). A survey on deep learning in medical image analysis. *Medical Image Analysis, 42*, 60-88.

Maas, A. L., Hannun, A. Y., & Ng, A. Y. (2013). Rectifier Nonlinearities Improve Neural Network Acoustic Models. *Proceedings of the 30th International Conference on Machine Learning*. Atlanta, Georgia, USA: JMLR: W&CP.

Macaluso, A., Clissa, L., Lodi, S., & Sartori, C. (2020). A Variational Algorithm for Quantum Neural Networks. *ICCS 2020: Computational Science Lecture Notes in Computer Science*, 12142, 591-604. Springer.

Nair, V., & Hinton, G. (2010). Rectified Linear Units Improve Restricted Boltzmann Machines. *ICML*, Omnipress, 807-814.

Nielsen, M. A., & Chuang, s. (2002). Quantum Computation and Quantum Information. *American Journal of Physics*, 70, 558.

Pedamonti, D. (2018). Comparison of non-linear activation functions for deep neural networks on MNIST classification task. *arXiv:1804.02763*.

Pereira, C. R., Pereira, D. R., Silva, F. A., Masieiro, J. P., Weber, S. A., Hook, C., & Papa, J. P. (2016a). A new computer vision-based approach to aid the diagnosis of Parkinson's disease. *Computer Methods and Programs in Biomedicine*, 136, 79-88.

Pereira, C. R., Weber, S. A., Hook, C., Rosa, G. H., & Papa, J. P. (2016b). Deep learning-aided Parkinson's disease diagnosis from handwritten dynamics. In *2016 29th SIBGRAPI Conference on Graphics, Patterns and Images (SIBGRAPI)*, 340-346. IEEE.

L. PARISI, D. NEAGU, R. MA, F. CAMPEAN – QRELU AND M-QRELU IN TENSORFLOW AND KERAS   30Pereira, C. R., Pereira, D. R., Papa, J. P., Rosa, G. H., & Yang, X. S. (2016c). Convolutional neural networks applied for Parkinson's disease identification. In *Machine learning for health informatics*, 377-390. Springer, Cham.

Qiu, S., Xu, X., & Cai, B. (2018). FReLU: Flexible Rectified Linear Units for Improving Convolutional Neural Networks. *arXiv*:1706.08098v2.

Rumelhart, D. E., Hinton, G. E., & Williams, R. J. (1986). Learning representations by back-propagating errors. *Nature*, 323, 533–536.

Russakovsky, O., Deng, J., Su, H., Krause, J., Satheesh, S., Ma, S., Fei-Fei, L. (2015). ImageNet Large Scale Visual Recognition Challenge. *International Journal of Computer Vision*, 115, 211-252.

Sakar, B. E., Isenkul, M. E., Sakar, C. O., Sertbas, A., Gurgen, F., Delil, S., ... & Kursun, O. (2013). Collection and analysis of a Parkinson speech dataset with multiple types of sound recordings. *IEEE Journal of Biomedical and Health Informatics*, *17*(4), 828-834.

Schuld, M., Sinayskiy, I., & Petruccione, F. (2014). The quest for a Quantum Neural Network. *Quantum Inf Process*, 13, 2567-2586.

Shang, W., Sohn, K., Almeida, D., & Lee, H. (2016). Understanding and improving convolutional neural networks via concatenated rectified linear units. *Proceedings of the 33rd International Conference on Machine Learning, JMLR.* NY, USA: W&CP.

Shi, H., Han, X., Jiang, N., Cao, Y., Alwalid, O., Gu, J., Zheng, C. (2020). Radiological findings from 81 patients with COVID-19 pneumonia in Wuhan, China: a descriptive study. *The Lancet*. Infectious diseases.

Siddique, F., Sakib, S., & Siddique, M. A. B. (2019, September). Recognition of handwritten digit using convolutional neural network in python with tensorflow and comparison of performance for various hidden layers. In 2019 *5th International Conference on Advances in Electrical Engineering (ICAEE)* (pp. 541-546). IEEE.

Solenov, D., Brieler, J., & Scherrer, J. F. (2018). The Potential of Quantum Computing and Machine Learning to Advance Clinical Research and Change the Practice of Medicine. *Mo Med.*, 115(5): 463–467.

TensorFlow. 2020. Tf.Nn.Leaky_Relu | Tensorflow Core V2.3.0. [online] Available at: <https://www.tensorflow.org/api_docs/python/tf/nn/leaky_relu> [Accessed 30 July 2020].

TensorFlow. 2020. Tf.Keras.Layers.Leakyrelu | Tensorflow Core V2.3.0. [online] Available at: <https://www.tensorflow.org/api_docs/python/tf/keras/layers/LeakyReLU> [Accessed 30 July 2020].

Verity, R., Okell, L. C., Dorigatti, I., Winskill, P., Whittaker, C., Imai, N., Cori, A. (2020). Estimates of the severity of coronavirus disease 2019: a model-based analysis. *Lancet*, 20 (6); 669-677.

WHO (Mar 2020). Retrieved from https://www.who.int/dg/speeches/detail/who-director-general-s-opening-remarks-at-the-media-briefing-on-covid-19---11-march-2020.

Xu, B., Wang, N., Chen, T., & Li, M. (2015). Empirical evaluation of rectified activations in convolutional network. *arXiv*:1505.00853.

Zeiler, M. D., & Fergus, R. (2014). Visualizing and Understanding Convolutional Networks. *ECCV 2014: European Conference on Computer Vision*, 818-833.

Zham, P., Kumar, D. K., Dabnichki, P., Poosapadi Arjunan, S., & Raghav, S. (2017). Distinguishing different stages of Parkinson's disease using composite index of speed and pen-pressure of sketching a spiral. *Frontiers in neurology*, *8*, 435.

Zhou, F., Yu, T., Du, R., Fan, G., Liu, Y., Liu, Z., Cao, B. (2020). Clinical course and risk factors for mortality of adult inpatients with COVID-19 in Wuhan, China: a retrospective cohort study. *Lancet*, 395(10229), 1054-1062.